\documentclass{article}
\usepackage{spconf,amsmath,epsfig}
\usepackage{amssymb}
\usepackage{amsthm}
\usepackage{algorithm}
\usepackage{algpseudocode}
\usepackage{tikz}
\usepackage{graphicx}
\usepackage{subcaption}
\usepackage{epstopdf}
\usepackage{epsfig}
\usetikzlibrary{tikzmark,calc}
\usepackage{caption}
\usepackage{float}
\usepackage{multirow}
\usepackage{todonotes}
\usepackage{microtype}
\usepackage{hyperref}
\usepackage{mathtools}
\usepackage{pgfplots}
\usepackage{rotating}
\usepackage{flushend} 
\usepackage{spconf,amsmath,graphicx,enumitem}
\usepackage{epstopdf}
\usepackage[font=footnotesize,labelfont=bf]{caption}
\usepackage{color,mathtools,amssymb}
\usepackage{balance}

\graphicspath{{figs/}}

\DeclareMathOperator*{\argmin}{arg\,min}

\def\R{{\mathbb{R}}}
\def\x{\mathbf{x}}
\def\R{{\mathbb{R}}}

\newcommand{\norm}[1]{\|#1\|}

\usepackage{graphicx}
\usepackage{amsmath,amsfonts,amsthm,amssymb,xspace,bm, verbatim,dsfont}
\usepackage{url,algorithm,algpseudocode}
\usepackage{thmtools}
\usepackage{thm-restate}
\usepackage{cleveref}
\usepackage{tikz}
\usetikzlibrary{decorations.pathreplacing,angles,quotes}
\usetikzlibrary{positioning,plotmarks,calc}
\usepackage{bbm}
\usepackage{todonotes}
\usepackage{pgfplots}
  
\theoremstyle{definition}
\newcommand{\note}[1]{\marginpar{\tiny *note in TeX*}}

\newcommand{\reals}{\mathbb{R}}

\newcommand{\vect}[1]{\mathbf{#1}}
\newcommand{\mat}[1]{\mathbf{#1}}

\newcommand{\iprod}[2]{\left\langle #1, #2 \right\rangle}

\newcommand{\twonorm}[1]{\left\| {#1} \right\|_2}

\newcommand{\sign}[1]{\operatorname{sign}\left(#1\right)}

\newcommand{\distop}[2]{\mathrm{dist}\left(#1,#2\right)}

\newcommand{\rbrak}[1]{\left(#1\right)}

\newcommand{\cbrak}[1]{\left\{#1\right\}}

\newcommand{\y}{\vect{y}}
\newcommand{\e}{\vect{e}}
\newcommand{\z}{\vect{z}}
\newcommand{\w}{\vect{w}}

\newcommand{\xo}{\vect{x^*}}

\newcommand{\ai}{\vect{a}_i}

\newcommand{\p}{\vect{p}}

\newcommand{\A}{\mat{A}}

\newcommand{\subspaces}{{\mathcal{M}}}

\newcommand{\mA}{\ensuremath{\mathbf A}}

\title{Alternating Phase Projected Gradient Descent with Generative Priors for Solving Compressive Phase Retrieval}

\name{Rakib Hyder$^1$, Viraj Shah$^2$, Chinmay Hegde$^{2}$, and M. Salman Asif$^{1}$
\thanks{RH and MA were supported by the DARPA REVEAL Program and an equipment donation from NVIDIA Corporation. VS and CH were supported in part by grants from NSF CCF-1815101, a Faculty Fellowship from the Black and Veatch Foundation, and an equipment donation from NVIDIA Corporation.}
}
\address{$^{1}$ Electrical and Computer Engineering , University of California, Riverside \\
$^{2}$ Electrical and Computer Engineering, Iowa State University}
\begin{document}
\ninept
\maketitle
\begin{abstract}
The classical problem of phase retrieval arises in various signal acquisition systems. Due to the ill-posed nature of the problem, the solution requires assumptions on the structure of the signal. In the last several years, sparsity and support-based priors have been leveraged successfully to solve this problem. In this work, we propose replacing the sparsity/support priors with generative priors and propose two algorithms to solve the phase retrieval problem. Our proposed algorithms combine the ideas from AltMin approach for non-convex sparse phase retrieval and projected gradient descent approach for solving linear inverse problems using generative priors. We empirically show that the performance of our method with projected gradient descent is superior to the existing approach for solving phase retrieval under generative priors. We support our method with an analysis of sample complexity with Gaussian measurements.

\end{abstract}

\section{Introduction}
\label{sec:intro}

\subsection{Motivation}

The classical problem of phase retrieval arises in numerous imaging applications \cite{shechtman2015phase, maiden2009improved}, where only the magnitude of the light rays can be measured but not the phase. As each linear observation loses its phase, the highly non-linear forward model makes it challenging to recover the underlying signal. The phase retrieval problem seeks to recover a real- or complex-valued unknown signal $\x^∗ \in \mathbb{R}^n$ from its (possibly noisy) amplitude-only observations $
\mathbf{y} \in \mathbb{R}^m$ of the form:
\begin{equation}
y_i= |\iprod{\ai}{\x^*}| + e_i, \;\;\;i=1,...,m,   
\label{eqn:magnitude-measurements}
\end{equation}
We construct $\mathbf{A} = \left[\mathbf{a_1~a_2~...~a_m}\right]^T$ with i.i.d. Gaussian entries. For simplicity, we ignore the noise $e_i$. We consider a setting with $m<n$, thus in general, the inverse problem in \eqref{eqn:magnitude-measurements} is highly ill-posed. 

In general, infinitely many possible solutions exist for \eqref{eqn:magnitude-measurements}. A conventional approach for solving such a problem is by constraining the solution to a set $\mathcal{M} \subseteq \R^n$ that captures some sort of known structure that $\x^*$ is expected to obey. The resulting optimization can be written as 
\begin{align}
\widehat{\mathbf{x}} &= \argmin~f(\y; |\mA \x|)~~\label{eq:cop}\\
&\text{s.t.}~~~\x \in \mathcal{M}.\nonumber
\end{align}
A common modeling assumption on $\x^*$ is \emph{sparsity}, which alleviates the ill-posed nature of the inverse problem, and in fact, makes the accurate recovery of $\x^*$ possible.

However, while being powerful from a computational standpoint, the sparsity prior has somewhat limited discriminatory capability, and it is certainly true that nature exhibits far richer nonlinear structure than sparsity alone. Thus, we focus on a newly emerging family of priors that are \emph{learned} from massive amounts of training data using generative networks such as GAN~\cite{goodfellow2014generative}. A well-trained generator closely captures the notion of a signal (or image) being `natural'~\cite{berthelot2017began}. While such generative priors have been used successfully in solving compressive sensing and other inverse problems~\cite{bora2017compressed}, including phase retrieval~\cite{hand2018phase,shamshad2018robust}, the optimal way to search for the solution within the range of generative prior has not yet been understood well. Most of these methods rely on loss minimization through gradient descent that often fails to search the entire solution space resulting in sub-optimal results. In this work, we provide two algorithms that enable an improved way of searching the solution space. Our work improves on the results of~\cite{hand2018phase,shamshad2018robust} empirically, along with providing mathematical analysis of convergence.
\begin{figure*}
	\centering
	\includegraphics[width=0.8\textwidth]{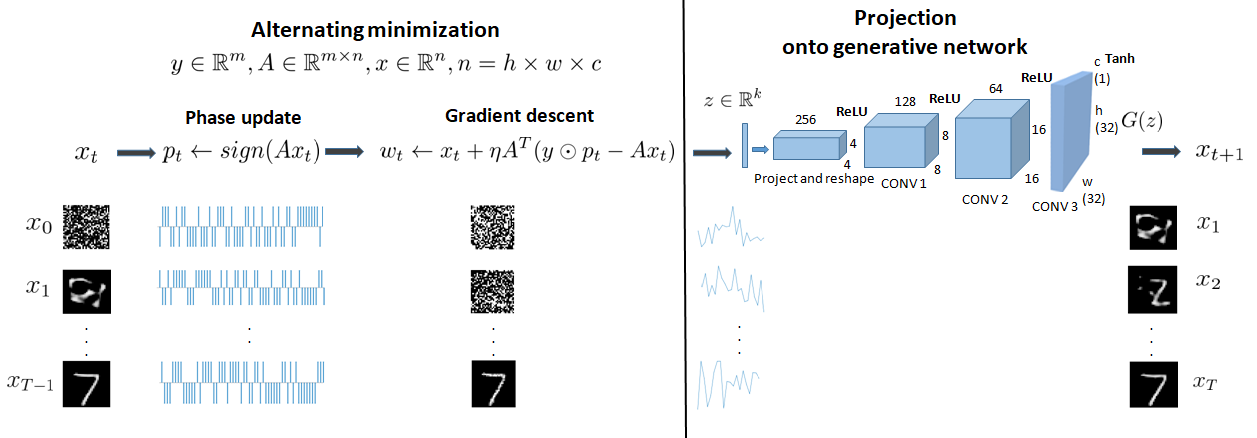}
	\caption{\emph{Illustration of APPGD algorithm. It has two major steps: alternating minimization and projection onto the range of the generator nework. In alternating minimization step, we update phase and perform one gradient descent update using the updated phase. Starting from a random vector, we perform phase update, gradient descent update step  and projection step iteratively to reach the final estimate.}\label{fig:intro1}}
\end{figure*}

\subsection{Our contributions}
In this paper, we propose and analyze two phase retrieval algorithms: alternating phase gradient descent (APGD), and alternating phase projected gradient descent (APPGD) to leverage generative priors. We improve over the approaches of ~\cite{hand2018phase,shamshad2018robust} by combining the gradient descent and projected gradient descent methods for generative priors~\cite{bora2017compressed,shah2018solving} with AltMin-based non-convex optimization techniques used in sparse phase retrieval~\cite{netrapalli2013phase,Jagatap2017}. 

We adopt a setting similar to~\cite{hand2018phase,shamshad2018robust}, and assume that the generator network (say, $G$) well approximates the high-dimensional
probability distribution of the set $\mathcal{M}$, \textit{i.e.}, we expect that for each vector $\x^*$ in $\mathcal{M}$, there exists a vector $\x = G(\z)$ that is very close to $\x^*$ in the support of the distribution defined by $G$. 
\[
\mathcal{M}=\{\x\in \mathbb{R}^n | \x=G(\z) \; \text{for some } \z \in \mathbb{R}^k\},
\]
With this assumption, the solution to \eqref{eq:cop} can be obtained by solving the following optimization problem: 
\begin{align}
    \widehat{\x} &= \argmin_\x\|\y-|\A\x|\|^2~~\label{eq:prob2}\\
    &~~~~~~~~~\text{s.t.}~~~\x = G(\z), \nonumber
\end{align}
where $\z$ is the latent code corresponding to image $\x$. Unless otherwise stated, all norms represented by $\|\cdot\|$ in this paper are Euclidean norms.

Recent work in \cite{hand2018phase,shamshad2018robust} minimizes the objective in \eqref{eq:prob2} directly over the latent variable $\z$ using gradient descent, and sets $\widehat{\x}$ as:
\begin{align}
    \widehat{\x} &= G(\argmin_\z\|\y-|\A G(\z)|\|^2).~~\label{eq:prob3}
\end{align}
 We refer to this approach as the ``gradient descent approach". Given that the generative models usually exhibit highly non-linear behavior, the above objective is highly non-convex. Moreover, direct application of gradient descent over $\z$ limits the explorable solution space, as at any stage it is not possible to explore the region outside the range of the generator. If initialized incorrectly, gradient descent can get stuck in local minima. In practice such algorithms require several restarts in order to provide good performance.
 
In phase retrieval problems, knowledge of phase and the signal is interdependent, as given the phaseless measurements just knowing the phase often enables us to estimate the signal. Thus, as alternative for solving \eqref{eq:prob2}, we can convert the phase retrieval problem to a linear inverse problem by initializing with a random phase $\p$ and update the phase with the solution of the linear inverse problem. Equation \eqref{eq:prob5} describes this approach for the $t^{th}$ iteration.
\begin{equation}
\begin{aligned}
    &\widehat{\x}_{t+1} = G(\argmin_\z\|\p_t\odot \y-\A G(\z)\|^2)~~\label{eq:prob5}\\
\end{aligned}
\end{equation}
We refer this approach as the alternating phase gradient descent (APGD) approach.
 
We also propose a third approach, in which we use projected gradient descent (PGD) to solve~\eqref{eq:prob5} directly in the ambient space based on~\cite{shah2018solving}. Through iterative projections, we are able to mitigate the effects of local minima and are able to explore the space outside the range of the generator (G). In PGD, we update our estimate of $\x$ with the standard gradient descent update rule, followed by projection of the output onto the span of generator, $G$. We refer this approach as the alternating phase projected gradient descent (APPGD) approach. 
 We provide theoretical analysis of our methods, along with extensive experimental results.

\subsection{Prior work}
\label{sec:prior}
Approaches for solving the phase retrieval problem can be broadly classified into  convex and non-convex approaches. Convex approaches usually consist of solving a constraint optimization problem after linearizing the problem. The PhaseLift algorithm \cite{candes2013phaselift} and its variations \cite{gross2017improved}, \cite{candes2015phasediff} come under this category. Typical non-convex approaches include approaches based on Amplitude flow \cite{wang2016sparse,wang2016solving} and Wirtinger flow \cite{candes2015phase, zhang2016reshaped,  chen2015solving, cai2016optimal}.

In recent work, phase retrieval for the cases where underlying signal is sparse is of growing interest. Some of the convex approaches for sparse phase retrieval include \cite{ohlsson2012cprl, li2013sparse,bahmani2015efficient,jaganathan2012recovery}. Similarly, non-convex approaches for sparse phase retrieval includes \cite{netrapalli2013phase, cai2016optimal, wang2016sparse}. Our alternating minimization (AltMin)-based approach in this paper is mainly inspired from the non-convex sparse phase retrieval framework advocated in \cite{netrapalli2013phase,Jagatap2017}. 

Different plug-and-play priors \cite{venkatakrishnan2013plug,tirer2019image,zhang2017learning} have also been proposed to leverage the effect of off-the-shelf image denoisers for solving inverse problems. Recently, various researchers have explored the idea of replacing the sparsity priors with generative priors for solving inverse problems. \cite{bora2017compressed, shah2018solving} provided gradient descent and PGD algorithms respectively to solve compressive sensing problem. We use these approaches for the signal estimation in our algorithm. \cite{hand2018phase,shamshad2018robust} solves the phase retrieval problem using generative priors through enforcing the prior directly by minimizing an empirical risk objective over the domain of the generator. In this paper, we improve over their idea by providing alternative approaches based on AltMin and projected gradient descent. 

\section{Algorithm}
\label{sec:alg}
In this section we describe the APPGD approach in details. At first, we train a generator $G : \R^k \rightarrow \R^n$ that maps a latent vector $\z \in \R^k$ to a high dimensional sample space $G(\z) \in \R^n$. We assume that our generator network can closely approximate the probability distribution of the set of natural images, $\mathcal{M}$ to which our original images $\x$ belong. With this assumption, we can limit our search for $\widehat{\x}$ only to the range of the generator function,  $\mathcal{M}$. The generator $G$ is assumed to be differentiable, and hence we use back-propagation for calculating the gradients of the loss functions involving $G$ for gradient descent updates.

In each iteration of the APPGD algorithm (Alg. \ref{alg:phase_gan}), three steps are performed: a phase update step, a gradient descent update step, and a projection step. 
\subsection{Phase update}
The first step is to calculate the phase of $\A\x$. For real $\A$ and $\x$, at the $t^{th}$ iteration, we update the phase estimate:
\[
\p_t = \text{phase}(\A\x_t) \coloneqq \sign{\A\x_t}.
\] 
After calculating the phase vector $\p$, we can use an element-wise product between $\p$ and $\y$ as an estimate of linear measurements and convert the phase retrieval problem into a linear inverse problem.

\subsection{Gradient descent update}
The second step is simply an application of a gradient descent update rule on the loss function $f(\cdot)$ which is given as:
\[
f(\x) \coloneqq \|\y\odot \p-\A\x\|^2.
\] 
Thus, the gradient descent update at the $t^{th}$ iteration is given by:
\[
\w_t \leftarrow \x_t + \eta \A^T(\y\odot \p_t-\A\x_t),
\] 
where $\eta$ is the learning rate.
\subsection{Projection step}
In projection step, we aim to find an image from the span of the generator, $\mathcal{M}$  which is closest to our current estimate $\w_t$. 
We define the projection operator $\mathcal{P}_G$ as follows:
\[
\mathcal{P}_G\left(\w_t\right) \coloneqq G\left(\argmin_{\z}L_{in}(\z)\right),
\]
where $L_{in}$ is the inner loss function defined as,
\[
L_{in}(\z) \coloneqq \|\w_t - G(\z)\|^2.
\]
We solve the inner optimization problem by running gradient descent with $T_{in}$ number of updates on $L_{in}(\z)$. The learning rate $\eta_{in}$ is chosen empirically for this inner optimization.

In each of the $T$ iterations, we run $T_{in}$ updates for calculating the projection. Therefore, $T \times T_{in}$ is the total number of gradient descent updates required in our approach.

\begin{algorithm}[t]
	\caption{\textsc{APPGD}}
	\label{alg:phase_gan}
	\begin{algorithmic}[1]
	\State \textbf{Inputs:} $\y$, $\A$, $G$, $T$, \textbf{Output:}  $\widehat{\x}$
	\State \text{Choose an initial point $\x_0 \in \mathbb{R}^n$} \hspace{17.8em} 
	\For {t = 1,\ldots T }
	\State $\p_{t-1}\leftarrow \sign{\A\x_{{t-1}}}$ 
	\State $\w_{t-1} \leftarrow \x_{t-1} + \eta \A^T(\y\odot \p_{t-1}-\A\x_{t-1})$ \hspace{8.8em} 
	\State $\x_{t} \leftarrow \mathcal{P}_G(\w_{t-1}) = G\left(\argmin_\z \|\w_{t-1} - G(\z)\|\right)$ \hspace{0.6em} 
	\EndFor
	\State $\widehat{\x} \leftarrow \x_{T}$
	\end{algorithmic}
\end{algorithm}

\subsection{Analysis} 
\label{subsec:descent}
This part of the algorithm is described in Lines 3-7 of Algorithm~\ref{alg:phase_gan}. We can prove that \emph{provided a good initial estimate ($\x_0$), the above algorithm (APPGD) provably converges to $\x^*$.}

The intuition is as follows. Ignoring the noise, the observation model in \eqref{eqn:magnitude-measurements} can be restated as follows:
\begin{align*}
\sign{\iprod{\ai}{\xo}}\odot y_i = \iprod{\ai}{\xo} ,
\end{align*}
for all $i=\{1,2,\ldots, m\}$. To ease notation, denote the \emph{phase vector} $\vect{p} \in \reals^m$ as a vector that contains the unknown signs of the measurements, i.e., ${p}_i = \sign{\iprod{\ai}{\x}}$ for all $i=\{1,2,\ldots,m\}$. Let $\vect{p}^*$ denote the true phase vector and let $\mathcal{P}$ denote the set of all phase vectors, i.e. $\mathcal{P} = \cbrak{\vect{p}:p_i = \pm 1, \forall i}$. Then our measurement model gets modified as:
\begin{align*}
\vect{p}^*\odot\y = \A \xo.
\end{align*}

Therefore, the recovery of $\xo$ can be posed as a (non-convex) optimization problem:
\begin{align} \label{eq:lossfunc}
\min_{\x \in \subspaces,\vect{p} \in \mathcal{P}} \norm{\A\x - \vect{p}\odot\y}^2
\end{align}

To solve this problem, we alternate between estimating $\vect{p}$ and $\x$.
We perform two estimation steps: 

\begin{enumerate}
\item if we fix the signal estimate $\x$, then the minimizer $\vect{p} \in \mathcal{P}$ is given in closed form as:
	\begin{align} \label{eq:phase_est}
	\vect{p}=\sign{\A\x} ,
	\end{align} 
\item and if we fix the phase vector $\vect{p}$, the signal vector $\x \in \subspaces$ can be obtained by solving:
	\begin{align} \label{eq:loss_min}
	\min_{\x \in \subspaces} \|\A \x-\vect{p}\odot \y\|_2 .
	\end{align}
\end{enumerate}

We now analyze our proposed descent scheme. We obtain:

\begin{restatable}{theorem}{convergence}
\label{thm:lin_convergence}
	Suppose we have an initialization $\x_0 \in \subspaces$ satisfying $\distop{\x_0}{\xo} \leq \delta_0 \twonorm{\xo}$, for $0 < \delta_0 < 1$, and suppose the number of (Gaussian) measurements,
	$$ m >  C \rbrak{k d \log n}, $$ 
	for some large enough constant $C$. 
	Then with high probability the iterates $\x_{t+1}$ of Algorithm~\ref{alg:phase_gan}, satisfy:
	\begin{align} \label{eq:mainconvergence}
	\distop{\x_{t+1}}{\xo} \leq {\rho}\, \distop{\x_{t}}{\xo},
	\end{align}
	where $\x_t,\x_{t+1}, \xo \in \mathcal{M}$, and $ 0 < \rho < 1$ is a constant.
\end{restatable}

\noindent{\textbf{Proof sketch:}} The high level idea behind the proof is that with a $\delta$-ball around the true signal $\xo$, the ``phase noise'' can be suitably bounded in terms of a constant times the signal estimation error. To be more precise, suppose that $\z^* = \A \xo = \p^*\odot\y$. Then, at any iteration $t$, we have:
\begin{align*}
\z_t &= \p_t \odot \y \\
&= \p^* \odot \y + (\p_t - \p^*) \odot \y\\
&= \z^* + \e_t,
\end{align*}
where $\e_t$ can be viewed as the ``phase noise''. Now, examining Line 6 of the above algorithm, we have that $\x_t$ is the output of APPGD after $t$ iterations. An ``unpacking'' argument similar to the one in \cite{Jagatap2017} indicates that:
\[
\norm{\x_t - \xo} \leq \alpha \norm{\x_{t-1} - \xo} + \beta \norm{\e_t} ,
\]
where $\alpha$ is a small enough constant. We will show that $\norm{\e_t}$ can be also bounded in terms of $\norm{\x_{t-1} - \xo}$, via Lemma \ref{lem:phase_err_bound} below. Consequently: 
\[
\norm{\x_t - \xo} \leq \rho\norm{\x_{t-1} - \xo},
\]
where $\rho$ is a small enough constant. 

We therefore achieve a per-step error reduction scheme if the initial estimate $x_0$ satisfies $\norm{x_0 - \x^*} \leq \delta_0 \norm{\x^*}$. This result can be trivially extended to the case where the initial estimate $x_0$ satisfies $\norm{x_0 + \x^*} \leq \delta_0 \norm{\x^*}$, hence giving the convergence criterion of the form (for $\rho < 1$): 
\begin{align*}
\distop{\x_{t}}{\xo} \leq \rho\, \distop{\x_{t-1}}{\xo}.
\end{align*}

We now state Lemma \ref{lem:phase_err_bound} without proof. A proof will be provided in an extended version of this paper. The proof is an adaptation of the seminal analysis of \cite{bora2017compressed}.

\begin{restatable}{lemma}{phaseerror} \label{lem:phase_err_bound}
	Suppose that the generator network model $G(\cdot)$ is comprised of $d$ layers of neurons with ReLU activation functions and weight matrices with bounded operator norms. As long as the initial estimate is a small distance away from the true signal $\xo \in \mathcal{M}$ ( i.e. $\distop{\x_0}{\xo} \leq \delta \norm{\xo}$)
	and subsequently,
	$\distop{x_t}{\xo} \leq \delta \norm{\xo}$, where $\x_t$ is the $t^{th}$ update of Algorithm~\ref{alg:phase_gan}, then the following bound holds for any $t \geq 0$:
	\begin{align*}
	\norm{\e_{t+1}} \leq  \rho_1 \norm{\x_t - \xo},
	\end{align*}
	with high probability, as long as $ m > C (kd \log n)$ and $\rho_1 < 1$ is a constant.
\end{restatable}
\vspace{-1.8em}

\begin{figure}
\centering
   	\begin{subfigure}[t]{0.30\textwidth}
		\includegraphics[width=\textwidth]{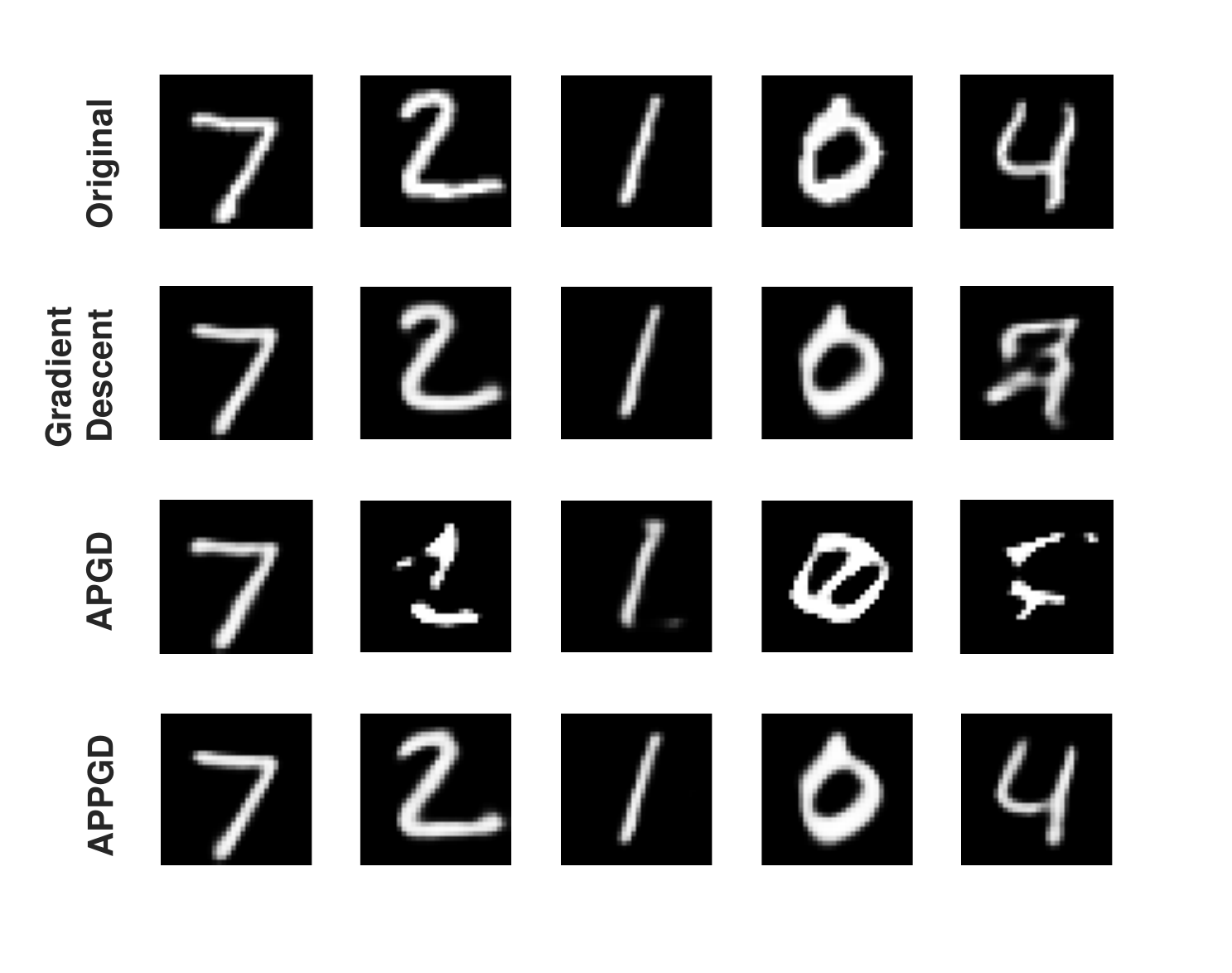}
		\caption{Reconstruction results on MNIST for three different  approaches with $m = 60$ measurements.}
		\label{fig:mnist-rec}
		\end{subfigure}\hfill%
	\begin{subfigure}[t]{0.23\textwidth}
		\includegraphics[width=\textwidth]{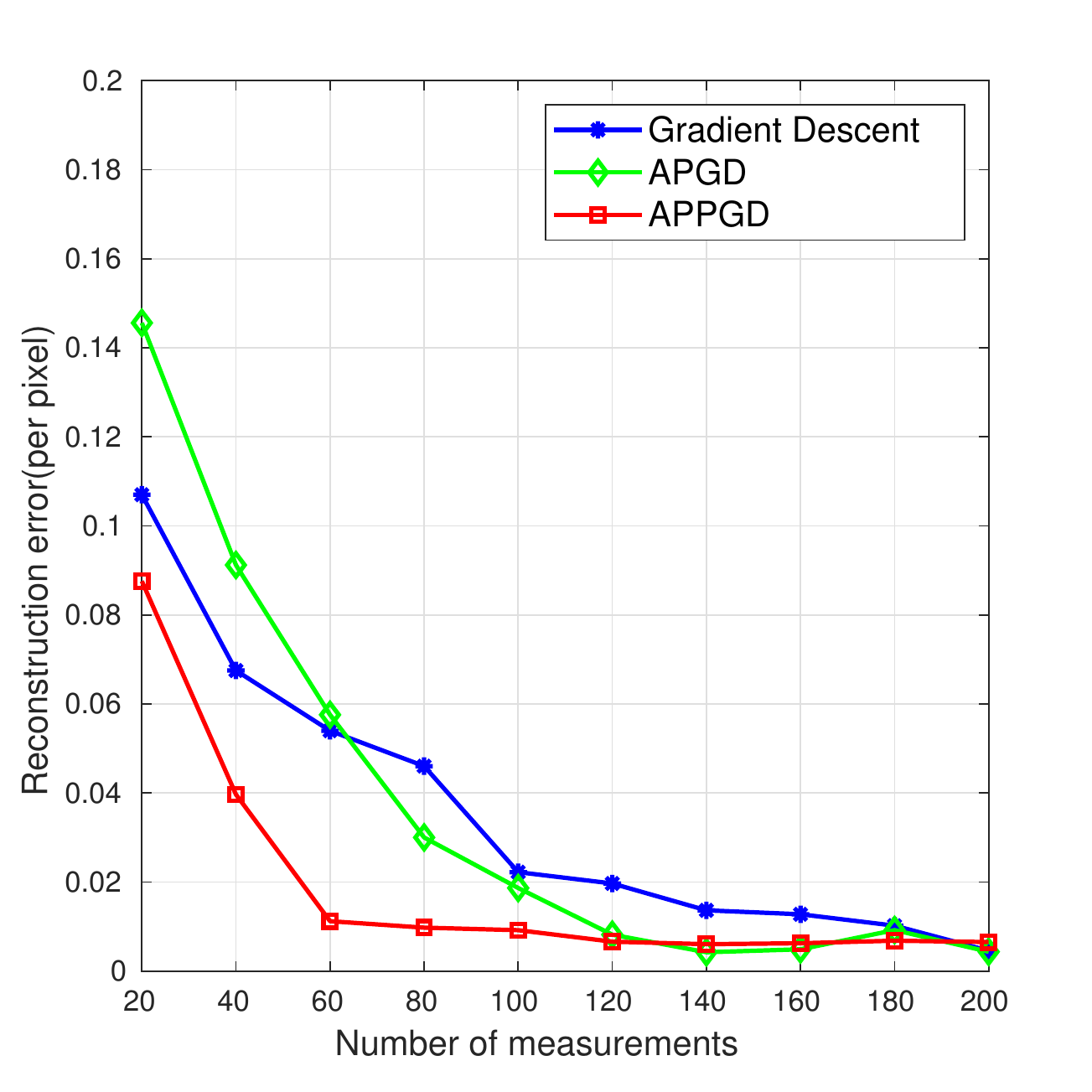}
		\caption{Reconstruction error (per pixel) for three approaches on MNIST.}
		\label{fig:mnist-mse}
	\end{subfigure}\hfill%
	\begin{subfigure}[t]{0.23\textwidth}
		\includegraphics[width=\textwidth]{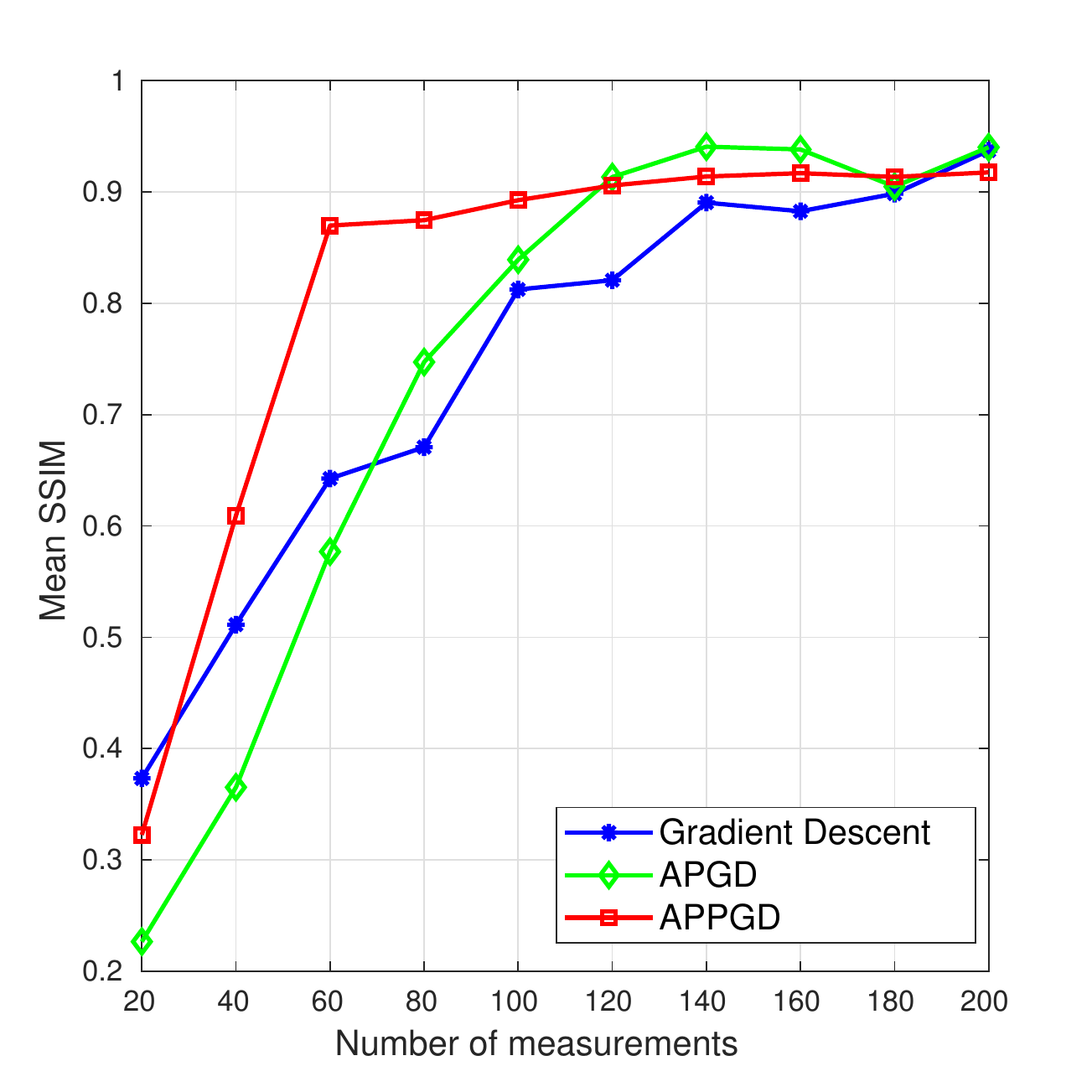}
		\caption{Mean SSIM for three approaches on MNIST.}
		\label{fig:mnist-ssim}
	\end{subfigure}
	\caption{\emph{Comparison of three approaches on MNIST test set.}}
\end{figure}

\begin{figure}
	\begin{subfigure}[t]{0.48\textwidth}
		\includegraphics[width=\textwidth]{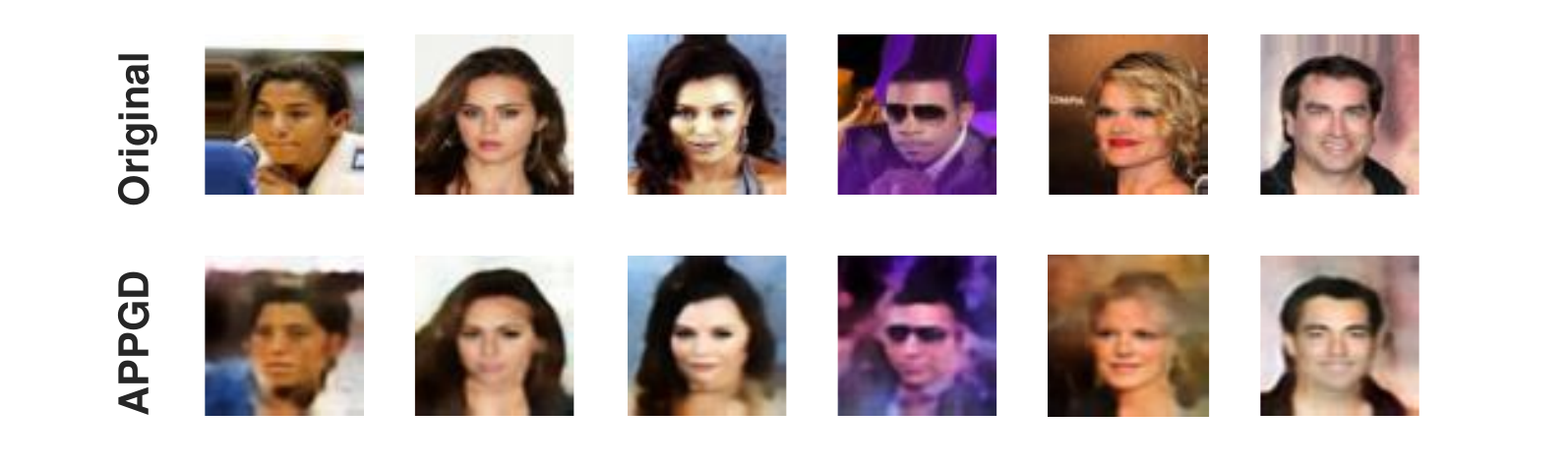}
		\caption{Reconstruction results on celebA dataset for APPGD with $m = 1000$ measurements.}
		\label{fig:celeba-rec}
		\end{subfigure}\hfill%
   	\begin{subfigure}[t]{0.23\textwidth}
		\includegraphics[width=\textwidth]{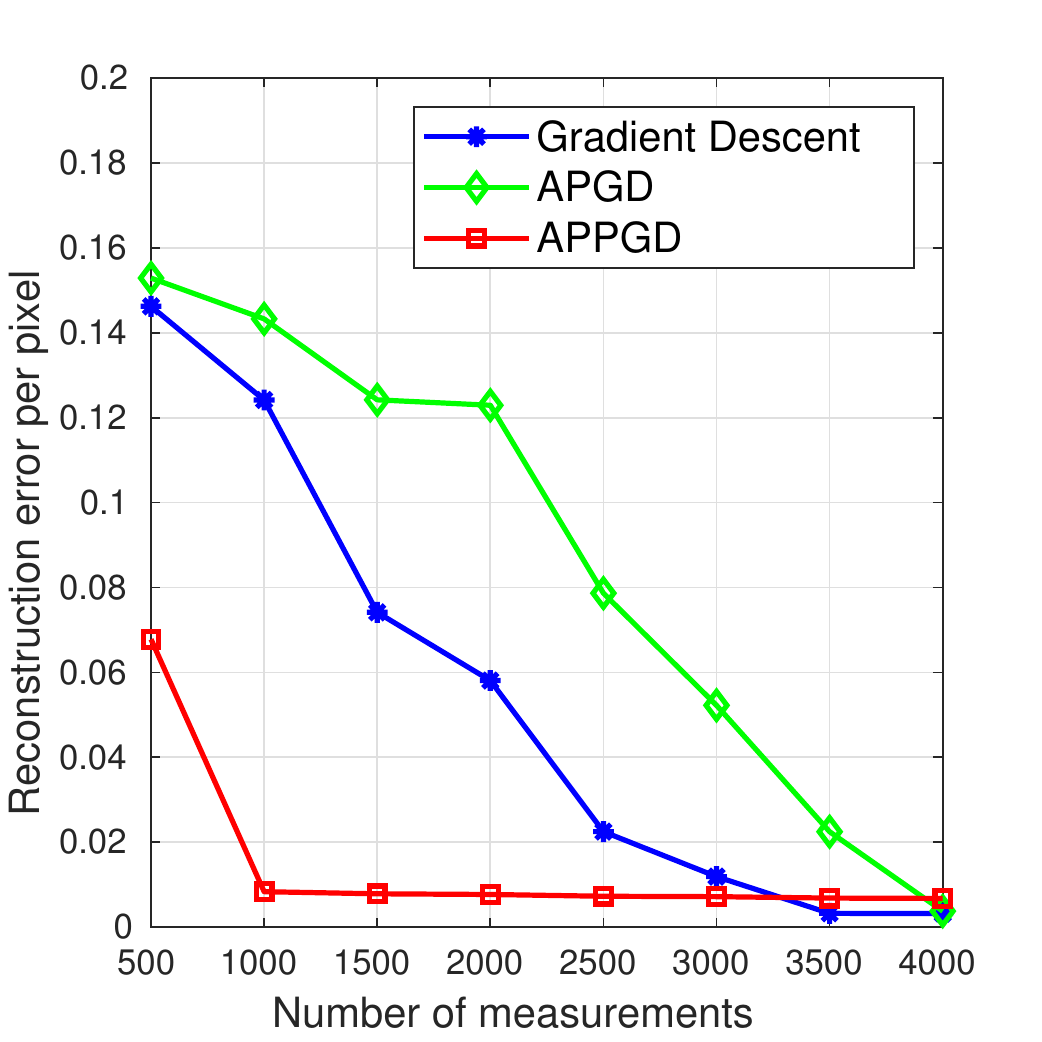}
		\caption{Reconstruction error (per pixel) for three approaches on celebA.}
		\label{fig:celeba-mse}
	\end{subfigure}\hfill%
	\begin{subfigure}[t]{0.23\textwidth}
		\includegraphics[width=\textwidth]{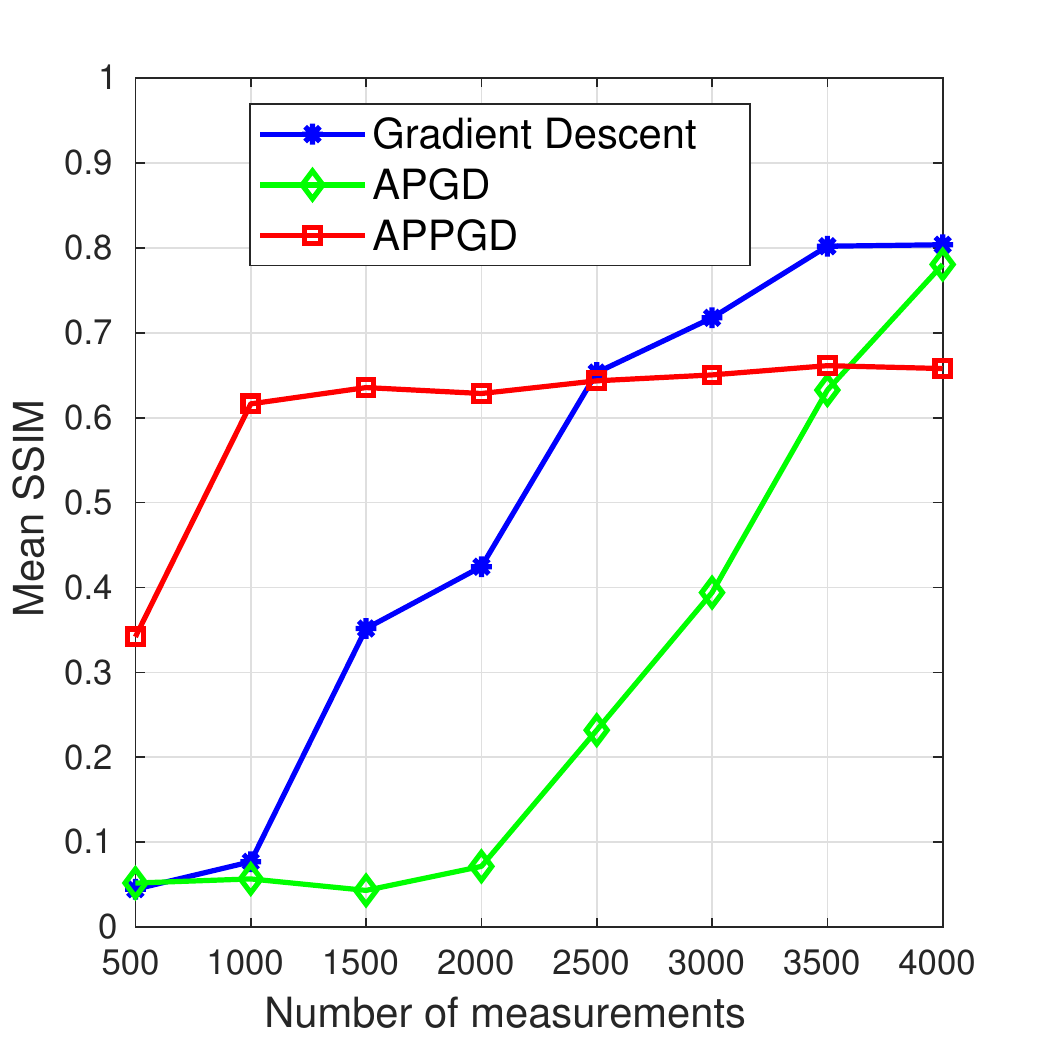}
		\caption{Mean SSIM for three approaches on celebA.}
		\label{fig:celeba-ssim}
	\end{subfigure}
	
	\caption{\emph{Comparison of three approaches on celebA test set and some reconstruction results for our APPGD algorithm.}}
\end{figure}

\section{Models and Experiments}
\label{sec:exp}

In this section, we describe our experimental setup and report the performance comparisons of the three approaches. We use two different generative models for the MNIST and CelebA datasets. The generative model for CelebA follows the DCGAN framework \cite{radford2015unsupervised} except that we do not use any batchnorm layer since the gradient for this layer is dependent on batch size and the distribution of the batch. The  generator architecture for MNIST experiments is shown in Fig. \ref{fig:intro1}. We train our generators by jointly optimizing generator parameters, $\gamma$ and the latent code, $\z$ using SGD optimization by following the procedure from \cite{Bojanowski2018OptimizingTL}. We use the squared-loss function, $l_2 (\x,\widehat{\x})=\|\x-\widehat{\x} \|^2 $ to train the generators. We choose $z$ from the standard normal distribution on $\mathbb{R}^k$ and then rescale it by its Euclidean norm. We project $\z$ back to the unit norm ball after each gradient update. 

In our experiments, we choose the entries of the matrix $A$ independently from the $\mathcal{N}(0,\frac{1}{m})$ distribution. Although we ignore the presence of noise, it is possible to replicate our experiments with additive Gaussian noise. For all the approaches we kept the number of update steps fixed. We do not allow random restarts. For fair comparison, we initialize $\x$ with the same random vector for all approaches and perform the same sign correction as in \cite{hand2018phase} on them.

We have our first set of experiments with three different approaches on a generator trained over the MNIST training dataset resized to $32\times32$ pixel. Considering that the representation error is very small, we test three approaches on 10 images from the test set of MNIST dataset and provide both quantitative and qualitative results. For APPGD, at the gradient descent step we choose $\eta=0.9$ because we need a meaningful output before passing it to the projection step \cite{shah2018solving}. We can also perform gradient descent multiple times at the first iteration before projecting it onto the range of generator so that we can start from a good initial point. For all three approaches, we use learning rate $\eta_{in}=0.01$. We use $T=50$ and $T_{in}=500$ for APPGD and APGD approaches. For fair comparison, we use 2500 iterations for Gradient descent approach. We measure reconstruction error, $\|\widehat{\x}-\x^*\|^2$, and SSIM for comparison. In Fig. \ref{fig:mnist-mse}, we show the reconstruction error comparisons and in Fig. \ref{fig:mnist-ssim} we show SSIM comparisons for increasing values of number of measurements. As the input images are not chosen from the span of the generator itself, it is not possible to reach zero error. However, we observe from \ref{fig:mnist-mse} that APPGD can reach near zero error with only 60 measurements which is significantly less than the other two approaches.  Fig. \ref{fig:mnist-rec} depicts reconstruction results for some of the selected MNIST images for three approaches.

For our second set of experiments, we train a generator for the CelebA dataset. For training, we resize the celebA dataset composed of 202,599 colored images of celebrity faces to $64\times64\times3$ and kept $\frac{1}{32}$ of the images apart. We do not use the aligned and cropped version which includes only the faces in the images. 

We experiment on a subset of 10 images from the held out test dataset and report reconstruction results. We set the total number of updates to $1500$, with $T = 50$ and $T_{in}=300$ for APGD and APPGD approaches. Learning rates for APPGD are set as $\eta = 0.9$ and $\eta_{in}=0.3$. Learning rates for APGD and Gradient Descent approaches are set as $\eta_{in} = 0.003$ (tuned to their best performance) for a fixed total number of updates. Image reconstruction results from $m=1000$ measurements with APPGD algorithm are displayed in Fig. \ref{fig:celeba-rec}. We show comparison of three approaches in terms of reconstruction error in \ref{fig:celeba-mse} and in terms of SSIM in \ref{fig:celeba-ssim}.  We observe that APPGD can achieve good reconstruction with far fewer measurements than the other competing approaches.

{{
\footnotesize
\bibliographystyle{IEEEbib}
\bibliography{refs}
}}

\end{document}